\documentclass{ecai} 

\usepackage{latexsym}
\usepackage{amssymb}
\usepackage{amsmath}
\usepackage{amsthm}
\usepackage{dsfont}
\usepackage[ruled,linesnumbered]{algorithm2e}
\usepackage{booktabs}
\usepackage{makecell}
\usepackage{enumitem}
\usepackage{multirow}
\usepackage{float} 
\usepackage{subfigure}
\usepackage{graphicx}
\usepackage{caption}
\usepackage{adjustbox}
\usepackage{color}

\newtheorem{definition}{Definition}

\begin{document}


\begin{frontmatter}


\paperid{5957} 


\title{RANA: Robust Active Learning for Noisy Network Alignment}


\author[A,B]{\fnms{Yixuan}~\snm{Nan}}
\author[A,B]{\fnms{Xixun}~\snm{Lin}\thanks{Corresponding Author. Email: linxixun@iie.ac.cn.}}
\author[A,B]{\fnms{Yanmin}~\snm{Shang}} 
\author[A,B]{\fnms{Zhuofan}~\snm{Li}}
\author[A,B]{\fnms{Can}~\snm{Zhao}}
\author[A,B]{\fnms{Yanan}~\snm{Cao}}

\address[A]{Institute of Information Engineering, Chinese Academy of Sciences, Beijing, China}
\address[B]{School of Cyber Security, University of Chinese Academy of Sciences, Beijing, China}


\begin{abstract}
Network alignment has attracted widespread attention in various fields. However, most existing works mainly focus on the problem of label sparsity, while overlooking the issue of noise in network alignment, which can substantially undermine model performance. Such noise mainly includes structural noise from noisy edges and labeling noise caused by human-induced and process-driven errors. To address these problems, we propose RANA, a Robust Active learning framework for noisy Network Alignment. RANA effectively tackles both structure noise and label noise while addressing the sparsity of anchor link annotations, which can improve the robustness of network alignment models. Specifically, RANA introduces the proposed Noise-aware Selection Module and the Label Denoising Module to address structural noise and labeling noise, respectively. In the first module, we design a noise-aware maximization objective to select node pairs, incorporating a cleanliness score to address structural noise. In the second module, we propose a novel multi-source fusion denoising strategy that leverages model and twin node pairs labeling to provide more accurate labels for node pairs. Empirical results on three real-world datasets demonstrate that RANA outperforms state-of-the-art active learning-based methods in alignment accuracy. Our code is available at \url{https://github.com/YXNan0110/RANA}.
\end{abstract}

\end{frontmatter}


\section{Introduction}
Network alignment, also known as anchor link prediction, aims to establish cross-network mappings between nodes across multiple networks \citep{saxena2024survey, trung2020comparative, senette2024user}. It has attracted widespread attention in various real-world cross-domain network analysis tasks, such as social network analysis \citep{li2021deep, deng2013personalized}, protein molecular comparison \citep{singh2007pairwise, singh2008global}, and fraud detection \citep{van2015apate, zhang2021real}. Most existing network alignment models rely on supervised learning, which typically requires a large number of high-quality pre-labeled anchor links as training data to achieve satisfactory alignment accuracy. However, for complex real-world networks, obtaining a large number of high-quality labeled anchor links is not only time-consuming but also resource-intensive. By prioritizing the labeling of the most valuable samples, active learning \citep{li2020seal, huang2024cost, wu2024robust} avoids wasting effort on uninformative labeled data, thereby significantly reducing overall labeling cost. For example, Attent \citep{zhou2021attent} utilizes information gain as the criterion for node selection, enabling it to identify the most informative nodes that contribute most to improving alignment performance. As a result, active learning-based methods have become one of mainstream directions in network alignment.

Although expressive, these active learning-based methods largely ignore the issues posed by noisy scenarios: \textbf{1) Structural noise}: Real-world networks often contain missing edges or incorrect connections~\citep{kang2019robust} that distort the graph topology used to learn node embeddings. However, most existing methods assume clean network structures, leading to inaccurate representations and making them

\begin{figure}[h]
\centering
\includegraphics[width=0.9\columnwidth]{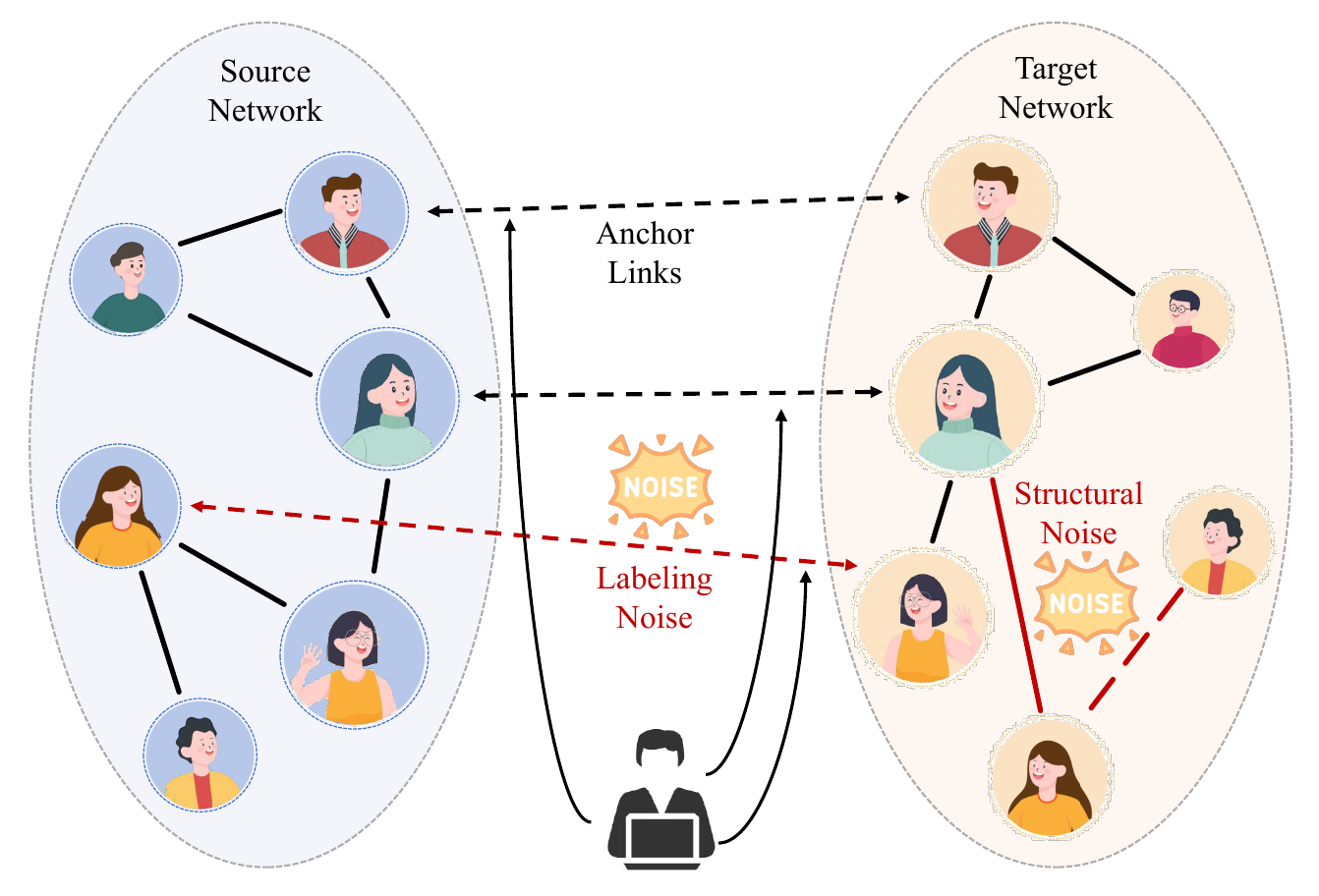}
\caption{An illustrative example of structural noise and labeling noise in network alignment.}
\vspace{1.5em}
\label{fig:figure5}
\end{figure}

\noindent
vulnerable to such noise, which can significantly degrade alignment performance. \textbf{2) Labeling noise}: The label quality completely depends on the manually annotated anchor link labels provided by the annotators (e.g., human annotators, crowdsourcing workers, or automated systems). The incorrect labeling of anchor links originates from two major sources: human-induced errors, such as annotators’ subjective judgments and limited domain knowledge, and process-driven errors arising from the labeling pipeline, including noisy data acquisition and inconsistencies in task execution, which in turn affect the alignment accuracy.

To address these critical issues as shown in Figure \ref{fig:figure5}, we propose a novel Robust Active learning framework for noisy Network Alignment\footnote{The description of noise network alignment is introduced in Section \ref{nna}.} (RANA), rethinking both sample selection and label trustworthiness under noise. RANA first aims to identify high-quality and informative node pairs, and then optimizing their labels to mitigate the impact of noise. Specifically, we introduce a \textbf{Noise-aware Selection Module}, which evaluates node pairs based on a combination of noise-aware confidence and node influence score. By incorporating a cleanliness score into confidence estimation, this module effectively prioritizes node pairs that are not only informative, but also less susceptible to structural noise. We further refine the labeling process through a \textbf{Label Denoising Module}, which addresses the issue of labeling noise. Rather than relying on a single source of annotation, this module employs an innovative multi-source fusion denoising strategy that integrates prediction information and similarity information to determine final labels. Our work provides more reliable labels, which can offer higher-quality inputs for network alignment model, ultimately enhancing its prediction performance. In general, our main contributions can be summarized as follows:
\begin{itemize}
    \item [1)]
    We formally define the problem of noisy network alignment, highlighting both structural and labeling noise introduce significant challenges to node alignment and sample selection, issues that have been largely overlooked in previous studies.
    \item [2)]
    We propose a robust active learning framework to address these two types of noise through two novel modules: the Noise-aware Selection Module, which optimizes the node pair selection by incorporating noise-aware confidence estimation, and the Label Denoising Module, which leverages model predictions and twin node pairs to provide high-quality labels.
    \item [3)]
    We validate the competitive performance of RANA on multiple public datasets. Notably, on the Facebook-Twitter dataset, our framework achieve a 6.24\% higher accuracy than baseline methods, demonstrating the effectiveness of our approach.
\end{itemize}

\vspace{-5pt}
\section{Related Work}

\subsection{Network Alignment}
Network alignment aims to identify nodes with the same identity across different networks, and it is widely applied in various fields, such as protein molecular recognition \citep{clark2015multiobjective, maskey2019survey, faisal2014global}, social network analysis \citep{yin2019social, zhao2023beyond, jiang2015social}, and fake information detection \citep{tan2014mapping, zhang2023capturing, rossi2015network}. Early network alignment studies~\citep{bartunov2012joint, zafarani2009connecting, koutra2013big} treated one network as a permutation of another, aligning them based on structural and attribute consistency constraints. For example, IsoRank \citep{singh2008global} performs alignment by analyzing the structural similarity of nodes between the source and target networks, employing a technique analogous to PageRank. FINAL \citep{zhang2016final} achieves network alignment by jointly enforcing topology, node attribute, and edge attribute consistency, while significantly reducing computational complexity via low-rank approximation. 

With the rise of deep learning and graph representation learning, increasing research has incorporated graph embedding into network alignment. PALE \citep{man2016predict} embeds each node to low-dimensional space to learn node representation, and uses MLP to learn node mapping between networks. GAlign \citep{trung2020adaptive} aligns node embeddings across all layers of GCN for capturing both global and local features and introduces a perturbation mechanism to enhance the model's robustness. 
NAME \citep{huynh2021network} models structural and attribute consistency using three distinct embeddings, and combines their alignment matrices through a reinforcement learning approach. PARROT \citep{zeng2023parrot}, grounded in optimal transport theory, introduces a position-aware transport cost and incorporates three consistency regularization terms to guide the alignment process. 
\par
However, the above methods struggle to handle structural and labeling noise effectively, leading to sub-optimal results in the setting of noisy network alignment.  

\subsection{Active Learning on Graphs}
Active learning is a machine learning paradigm in which a model proactively identifies and selects the most informative samples for labeling \citep{joshi2009multi, konyushkova2017learning}. By strategically querying samples, active learning can reduce the need for extensive manual annotation while maintaining high model performance. A key advantage of active learning lies in its ability to address label sparsity, a property that is particularly important in graph-structured data. Given the inherent sparsity of graphs\citep{glaria2021compact}, active learning naturally complements graph-based tasks, leading to its extension from traditional i.i.d. data settings to graph domains. AGE \citep{cai2017active} combines node centrality and node embeddings to select the most informative nodes as training samples. RIM \citep{zhang2021rim} introduces a reliable impact maximization strategy, considering both the quantity and quality of node influence. IGP \citep{zhang2022information} proposes a soft-label querying method, selecting nodes based on maximizing information gain propagation. \citet{fuchsgruber2024uncertainty} demonstrate that epistemic uncertainty can serve as an effective basis for active learning, showing that selecting nodes with the highest epistemic uncertainty can significantly boost the model’s confidence in predicting unlabeled nodes.

In network alignment, active learning is introduced to select nodes or node pairs, playing a crucial role in alleviating the problem of label sparsity. For example, \citet{malmi2017active} proposed two active learning methods, TopMatchings and GibbsMatchings, which use structural features, employing marginal distributions and cross-entropy to quantify determinism. DAULAP \citep{cheng2019deep} argues that anchor links are more valuable than non-anchor links and proposes three node pair selection strategies: cross-network information entropy, cosine similarity, and expected error reduction. Attent \citep{zhou2021attent} evaluates the quality of candidate query nodes using influence functions. 

Unlike previous works, RANA introduces noise-aware confidence, incorporating structural noise and labeling noise into confidence calculation. Additionally, our label denoising module enhances alignment accuracy while reducing the overall labeling budget.

\section{Preliminaries}
In network alignment, we focus on two networks: the source network $G^s=(V^s, E^s, \mathbf{A}^s)$ with node attribute matrix $\mathbf{X}^s$, and the target network $ G^t=(V^t, E^t, \mathbf{A}^t)$ with node attribute matrix $\mathbf{X}^t$. $V^s$ and $V^t$ represent the node sets, and $E^s$ and $E^t$ represent the edge sets. $N^s=|V^s|$, $N^t=|V^t|$. The source and target networks are related by a pre-alignment matrix $\mathbf{P}\in\{0, 1\}^{N^s\times N^t}$, where $[\mathbf{P}]_{ij}=1$ indicates that $v_i^s\in V^s$ and $v^t_j \in V^t$ is a pair of pre-aligned anchor nodes. In practice, the pre-alignment matrix is typically extremely sparse, providing only limited supervision, label sparsity emerges as a key challenge in network alignment. The network alignment model predicts anchor links, typically represented by an alignment matrix $\mathbf{S}$, where $[\mathbf{S}]_{uv}$ indicates the similarity between nodes $u \in V_s$ and $v \in V_t$. Formally, we define the network alignment problem as follows:
\begin{definition}
Network Alignment: Given two networks $G^s$ and $G^t$ with a pre-alignment matrix $\mathbf{P}$, the goal of network alignment is to calculate an alignment matrix $\mathbf{S}$ where $[\mathbf{S}]_{ij}>[\mathbf{S}]_{ir|r\ne j}$ iff there exists an anchor link between node $v^s_i$ and $v_j^t$.
\end{definition}

Given a pair of networks $G^s$ and $G^t$ with a pre-alignment matrix $\mathbf{P}$, active learning aims to improve the network alignment model by iteratively selecting and labeling informative node pairs. The process starts with a small set of initially labeled node pairs, used to train a preliminary model. At each iteration, the model selects the top $k$ most informative node pairs based on a sampling strategy. These pairs are then labeled by an oracle (e.g., an human annotator). The newly labeled data is added to the training set to update the model. Active learning is an iterative process in which increasingly more samples are labeled, leading to continuous improvements in model performance.

\vspace{-5pt}
\section{Method}

In this section, we first define the noisy network alignment. Then, we introduce the overview of RANA and its two modules, the Noise-aware Selection Module and the Label Denoising Module. 

\vspace{-10pt}
\subsection{Problem Definition}\label{nna}
\textbf{Noisy network alignment} includes both structural noise and labeling noise. Structural noise arises from missing or incorrect connections between nodes or edges, while labeling noise primarily arises from the subjective judgment of the annotator. Such types of noise are ubiquitous in real-world scenarios, making noisy network alignment an essential task.

Given a pair of noisy networks $G^s$ and $G^t$, a pre-alignment matrix $\mathbf{P}$, an oracle with labeling accuracy $\alpha$, and a query budget $k$, the objective of active learning in the context of noisy network alignment is to maximize the alignment accuracy of the network alignment model by selecting and labeling the most informative $k$ node pairs.

Similar to previous works~\citep{ren2019activeiter, zhou2021attent}, we set the oracle to answer the following question: Given a pair of nodes $(v_i^s, v_j^t)$ in the source network $G^s$ and the target network $G^t$, the oracle can judge whether there is an anchor link between the two nodes with accuracy $\alpha$.

\vspace{-5pt}
\subsection{Framework Overview}
\begin{figure*}[t!]
\captionsetup{justification=raggedright}
\includegraphics[width=1.0\textwidth]{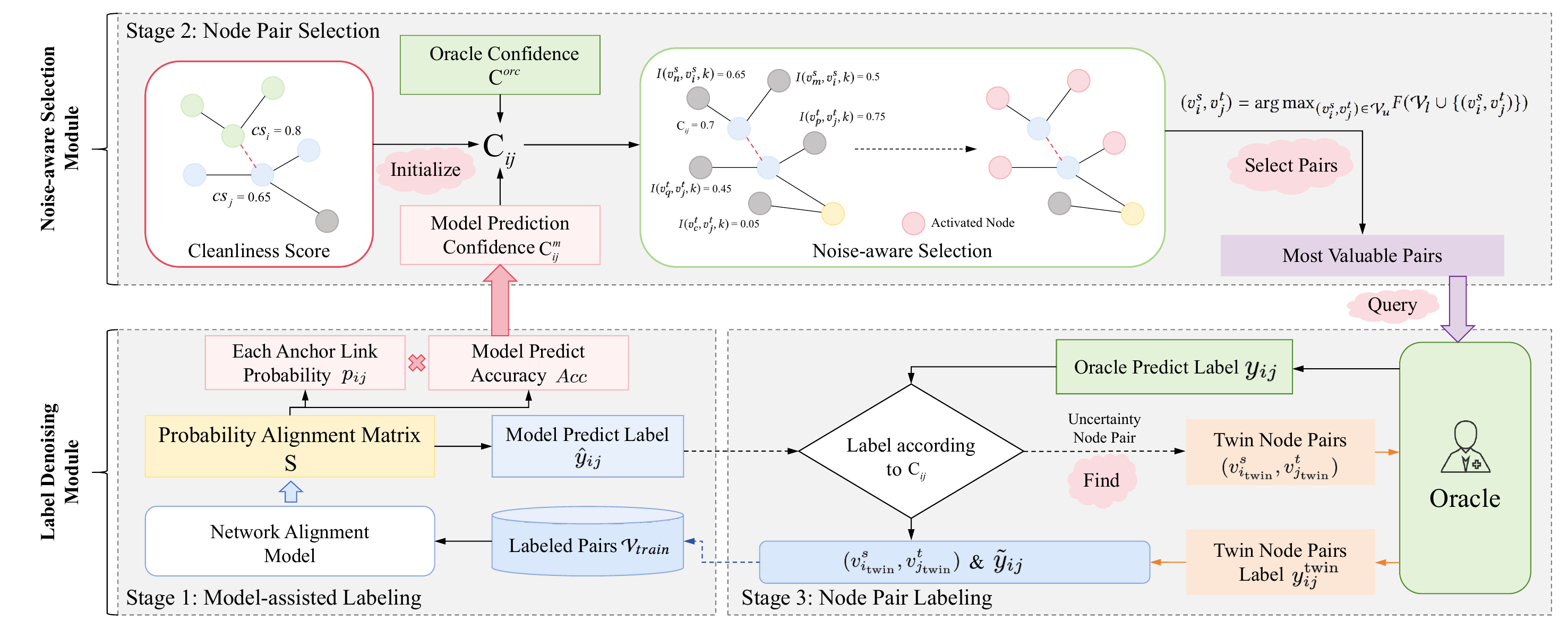}
\caption{\textbf{RANA Framework.} Stage 1 uses the trained model to generate prediction probability and estimate prediction accuracy, which are then passed to Stage 2. Stage 2 computes confidence scores and selects informative node pairs for Stage 3. Stage 3 assigns reliable labels to the selected node pairs and adds them to the labeled dataset for subsequent model training.}
\vspace{0.5em}
\label{fig:Figure 1}
\end{figure*}

As shown in Figure \ref{fig:Figure 1}, RANA mainly consists of three stages: model-assisted labeling, node pair selection, and node pair labeling. In Stage 1, RANA trains the network alignment model using labeled data and predicts anchor links, generating both the probability of each anchor link and the model’s prediction accuracy, which serve as guidance for the following process. In Stage 2, RANA selects node pairs by jointly considering noise-aware confidence and influence score, identifying node pairs with the highest noise-aware confidence and influence. In Stage 3, our multi-source fusion denoising strategy ensures label consistency and reduces errors caused by subjective judgment. The node pair selection stage is introduced in the following Noise-aware Selection Module. The model-assisted labeling stage and the node pair labeling stage are both labeling processes, so they are introduced in the Label Denoising Module.

\vspace{-5pt}
\subsection{Noise-aware Selection Module}
\textbf{Noise-aware Confidence.} As part of the selection criterion in our active learning framework, noise-aware confidence accounts for both the structural noise in the local region of node pairs and the labeling noise introduced during the labeling process.

To account for structural noise in the graph, we introduce a cleanliness score into our confidence, enabling the active learning process to prioritize node pairs with lower structural noise. This cleanliness score reflects the reliability of the structure based on feature similarity between adjacent nodes. Specifically, for a given node, its cleanliness score is defined as the average feature similarity (e.g., cosine similarity) between the node and its neighboring nodes. A higher cleanliness score indicates a more reliable neighborhood structure. The cleanliness score of node pair is then obtained by averaging the cleanliness scores of the source and target node. 
\begin{definition}
Cleanliness Score: For a node pair $(v_i^s, v_j^t)$, the cleanliness score $cs_{ij}$ is defined as
\begin{equation}
cs_{ij}=\frac{1}{2}(\sum_{v_m^s\in N(v_i^s)}\cos(\mathbf{x}_i^s, \mathbf{x}_m^s) +\sum_{v_n^t\in N(v_j^t)}\cos(\mathbf{x}_j^t, \mathbf{x}_n^t) ),
\end{equation}
where $N(v_i^s)$ represents the set of neighbors of node $v_i^s$, and $\cos(\mathbf{x}_i^s,\\ \mathbf{x}_m^s)$ uses the cosine function to measure the feature similarity between $v_i^s$ and $v_m^s$.    
\end{definition}

In addition to structural noise, labeling noise must also be considered in noise-aware confidence estimation. Since the annotation of node pairs is completed by noisy oracle and model-assisted labeling, it is essential to incorporate both the oracle confidence and the model prediction confidence into the calculation. We model the oracle confidence $\mathrm{C}^{orc}$ as its accuracy $\alpha$, typically derived from prior probabilities in practical settings.

As more anchor links are labeled, the network alignment model receives stronger supervision, resulting in improved prediction accuracy. However, even well-trained models may still misclassify particularly challenging node pairs. Hence, we estimate the overall model prediction confidence by combining the prediction accuracy of the model and the predicted probability of each node pair.
\begin{definition}
Model Prediction Confidence: Given the model prediction accuracy $Acc$ on the current training data and the model's predicted probability $p_{ij}$ for the node pair $(v_i^s, v_j^t)$, the model prediction confidence for the node pair $(v_i^s, v_j^t)$, denoted as $\mathrm{C}_{ij}^m$, is defined as
\begin{equation}
\mathrm{C}_{ij}^m=Acc\cdot p_{ij},
\end{equation}
where $p_{ij}$ is the prediction probability of the node pair, which is normalized from the alignment matrix predicted by the network alignment model.    
\end{definition}

The probability of anchor links and prediction accuracy, obtained during the model-assisted labeling stage, serve as prior knowledge in the selection stage. 

Combining the oracle confidence and the model prediction confidence, we further calculate the confidence for each node pair. The label for each node pair is determined jointly by the oracle prediction, the model's prediction, and the cleanliness score, as described in Section \ref{sec4.3}. Based on the prediction of node pair labels, we define the node pair confidence as follows.
\begin{definition}\label{eq4}
Node Pair Confidence: Given the oracle confidence $\mathrm{C}^{orc}$, the model prediction confidence $\mathrm{C}^m_{ij}$ for node pair $(v_i^s, v_j^t)$, and the node pair’s cleanliness score $cs_{ij}$, the confidence for node pair $(v_i^s, v_j^t)$, denoted as $\mathrm{C}_{ij}$, is defined as
\begin{itemize}
  \item \textbf{High-confidence region:} If $\mathrm{C}^m_{ij} \geq \mathrm{C}^{orc}$, then  
  \[
  \mathrm{C}_{ij} = \mathrm{C}^m_{ij};
  \]

  \item \textbf{Moderate-confidence region:} If $\mathrm{C}^{orc} > \mathrm{C}^m_{ij} > \gamma$, then
  \begin{itemize}
    \item If $y_{ij} = \hat{y}_{ij}$:
    \[
    \mathrm{C}_{ij} = 
    \frac{\mathrm{C}^{orc} \cdot \mathrm{C}^m_{ij}}
         {\mathrm{C}^{orc} \cdot \mathrm{C}^m_{ij} + (1 - \mathrm{C}^{orc}) \cdot (1 - \mathrm{C}^m_{ij})},
    \]

    \item If $y_{ij} \ne \hat{y}_{ij}$ and $y_{ij}^{\mathrm{twin}} = y_{ij}$:
    \[
    \mathrm{C}_{ij} = 
    \frac{\mathrm{C}^{orc} \cdot (1 - \mathrm{C}^m_{ij})}
         {1 - \mathrm{C}^{orc} \cdot \mathrm{C}^m_{ij}},
    \]

    \item If $y_{ij} \ne \hat{y}_{ij}$ and $y_{ij}^{\mathrm{twin}} = \hat{y}_{ij}$:
    \[
    \mathrm{C}_{ij} = 
    \frac{\mathrm{C}^m_{ij} \cdot (1 - \mathrm{C}^{orc})}
         {1 - \mathrm{C}^{orc} \cdot \mathrm{C}^m_{ij}};
    \]
  \end{itemize}

  \item \textbf{Low-confidence region:} If $\mathrm{C}^{orc} > \gamma > \mathrm{C}^m_{ij}$, then  
  \[
  \mathrm{C}_{ij} = \min(cs_{ij}, \mathrm{C}^{orc}),
  \]
\end{itemize}
where $y_{ij}$ represents the label assigned by the oracle for node pair $(v_i^s, v_j^t)$, $\hat{y}_{ij}$ represents the model prediction label for the node pair, and $y^{\mathrm{twin}}_{ij}$ represents the label assigned by the oracle for the twin node pair $(v_{i_{\mathrm{twin}}}^s, v_{j_{\mathrm{twin}}}^t)$ corresponding to $(v_i^s, v_j^t)$.
\end{definition}

The proof of Definition \ref{eq4} can be found in Appendix \ref{sec4.5}. When the model prediction confidence exceeds that of the oracle, the model's prediction is regarded as sufficiently trustworthy, eliminating the need for further annotation by the oracle. Conversely, if the model's confidence falls below the threshold $\gamma$, the prediction is deemed unreliable and excluded from further consideration. For cases where the model's confidence falls between the threshold and the oracle's accuracy, we calculate node pair's confidence by considering both the probability of correct predictions by the model and the accuracy of the oracle. The confidence of a node pair combines the predictions of both the oracle and the model, with higher confidence indicating more reliable labels for the node pair.

\textbf{Node Pair Selection.} The noise-aware confidence is designed to guide active learning toward selecting node pairs that are easier to label. Meanwhile, active learning should also prioritize the most valuable node pairs to achieve better predictive performance under a limited labeling budget. To this end, we introduce the influence score to estimate the value of node pairs and use the number of activated nodes as the final selection criterion.

\begin{definition}
Influence Score: The influence score of node $v_i$ on node $v_j$ after $k$-step propagation is the L1-norm of the expected Jacobian matrix $\hat{I}(v_j, v_i, k)=||\mathbb{E}[\partial{\mathbf{X}_j^{(k)}}/\partial{\mathbf{X}_i^{(0)}}]||_1$. Formally, the normalized influence score is defined as
\begin{equation}
I\left(v_{m}, v_{i}, k\right)=\frac{\hat{I}\left(v_{m}, v_{i}, k\right)}{\sum_{v_{w} \in V} I\left(v_{m}, v_{w}, k\right)} .
\end{equation}    
\end{definition}
The influence of node $v_i$ on $v_m$ is measured by the degree to which the change in the input features of node $v_i$ affects the aggregate features of node $v_m$ after $k$ propagation iterations \citep{wang2020unifying}. The influence score $I(v_m, v_i, k)$ captures the sum of the probabilities of all possible influence paths from node $v_i$ to $v_m$, which can be regarded a random walk process. The larger $k$ is, the harder it is to get a path from $v_i$ to $v_m$, and the influence score $I(v_m, v_i, k)$ will also decay as the distance $k$ increases. 

By jointly considering the noise-aware confidence and the influence score, we quantify confidence influence as the number of activated nodes as the node pair selection criterion. When the confidence influence of a node pair with respect to the target node exceeds the activation threshold $\theta$, the node pair is considered capable of activating the target node.
\vspace{-5pt}
\begin{definition}
Activated Nodes: Given the influence of a node pair on both ends $I(v, v_i^s, k)$, $I(u, v_j^t, k)$, the node pair confidence $\mathrm{C}_{ij}$, the activation threshold $\theta$, the set of nodes activated by the node pair $\sigma((v_i^s, v_j^t))$ is defined as
\begin{equation}
\sigma((v_i^s, v_j^t))=\bigcup_{\mathrm{C}_{ij} \cdot  I(v, v_i^s, k) \ge \theta}\{v\}\\
\bigcup_{\mathrm{C}_{ij} \cdot  I(u, v_j^m, k) \ge \theta}\{u\}.
\end{equation}
\end{definition}
The activation threshold $\theta$ is a hyperparameter. When $\theta=0$, the receptive field is maximized, and the number of potentially activated nodes is the greatest. Increasing the activation threshold slightly reduces the number of nodes that can be activated, but improves the "purity" of the activated nodes. The activated nodes have higher similarity and greater confidence, which better reflects the "effective influence" of the node pair, thereby allowing the selection of more valuable node pairs.

Inspired by maximizing the influence of social networks \citep{aral2018social}, the goal of active learning in network alignment is to activate more nodes, enabling the alignment model to learn more comprehensive alignment knowledge under a limited budget. Therefore, our objective is to select a subset $\mathcal{V}_l$ from the unlabeled node pairs $\mathcal{V}_u$ such that the number of activated nodes is maximized. The noise-aware maximization objective is defined as
\begin{equation}\label{objective}
\underset{\mathcal{V}_l}{\max} F(\mathcal{V}_l)=\bigcup_{(v_i^s, v_j^t)\in \mathcal{V}_l} |\sigma((v_i^s, v_j^t))|, \quad \mathbf{s.t.}\quad \mathcal{V}_l\in \mathcal{V}_u,|\mathcal{V}_l|=\mathcal{B}.
\end{equation}
This objective aims to identify node pairs with minimal noise and maximal influence, whose selection yields the greatest increase in the number of activated nodes across both the source and target networks. By selecting these node pairs, we maximize the alignment model’s learning potential within a limited labeling budget, ensuring that the model captures the most critical alignment information with minimal label costs.

\vspace{-8pt}
\subsection{Label Denoising Module}
\label{sec4.3}
The label denoising module employs a multi-source fusion selection strategy throughout its two-stage process, consisting of the model-assisted labeling stage and the node pair labeling stage. The model-assisted labeling stage is for denoising high-confidence node pairs, and the node pair labeling stage is for denoising uncertain node pairs.

\textbf{Model-assisted Labeling.} With the accumulation of labeled anchor links, the network alignment model improves in accuracy and eventually surpasses the noisy oracle, making it reliable enough to provide labels directly.

Given that the accuracy of the oracle is $\alpha$, the current prediction accuracy of the model is $Acc$, and the predicted probability that the node pair is an anchor link is $p_{ij}$, then the label predicted by the model is
\begin{equation}
\hat{y}_{ij}=f(\mathcal{V}_{train}), \quad Acc\cdot p_{ij}>\alpha.
\end{equation}
With continued training, not only does the model's accuracy increase, but the number of anchor links labeled by the model also grows. Compared to relying on the oracle, model-assisted labeling becomes a more efficient and cost-effective approach.

Because the oracle labels of node pairs are subject to a certain probability of mislabeling, we introduced model-assisted labeling in the previous section. When the accuracy of model-assisted labeling falls between the oracle’s accuracy and the minimum acceptable threshold $\gamma$, model's labels remain usable. Unlike the oracle's binary labels ("anchor link" or "no anchor link"), the model provides probability values, offering richer information. This is particularly valuable when the oracle's labels are unreliable, and the model can provide more accurate label predictions for the node pairs.

\textbf{Twin Node Pairs Labeling.} We propose a twin node pair labeling strategy to address uncertain node pairs—those for which the model’s predicted labels $\hat{y}_{ij}$ conflict with the oracle’s predictions$y_{ij}$. To resolve such inconsistencies, we select the most similar node pairs as twin node pairs and query the oracle for the labels of twin node pairs to determine the labels of the original node pairs. For graph learning tasks, the information in the graph includes both feature information and structural information. Both factors jointly determine the similarity between node pairs. Therefore, a twin node pair is defined as the one most similar to the original in terms of both structure and node features.
\begin{definition}
Twin Node Pairs: Given a node pair $(v_i^s, v_j^t)$, its twin node pair $(v_{i_{\mathrm{twin}}}^s, v_{j_{\mathrm{twin}}}^t)$ is defined as the most similar pair based on feature and structure similarity:
{\small
\begin{equation}\label{eq10}
(v_{i_{\mathrm{twin}}}^s, v_{j_{\mathrm{twin}}}^t)= \\
\underset{(v_{i_{\mathrm{twin}}}^s, v_{j_{\mathrm{twin}}}^t)}{\arg\min}(S(\hat{\mathbf{X}}_{i_{\mathrm{twin}}}^s, \hat{\mathbf{X}}_{i}^s)+S(\hat{\mathbf{X}}_{j_{\mathrm{twin}}}^t, \hat{\mathbf{X}}_{j}^t)),
\end{equation}
}
where $\hat{\mathbf{X}}^s = \mathbf{X}^s (\mathbf{A}^s)^2$ incorporates both node features and 2-hop structural information. $S(\cdot)$ represents the similarity calculation function between nodes, such as cosine similarity.
\end{definition}

After obtaining twin node pairs, we query the oracle to obtain the label $y_{ij}^{\mathrm{twin}}$ of twin node pair. The node pair label adopts the oracle's label $y_{ij}$ if consistent with its twin node pair, otherwise follows the model prediction $\hat{y}_{ij}$ upon twin consensus. Based on the twin node pair labeling mechanism, twin node pairs and the original node pairs exhibit high similarity, allowing us to obtain the labels of both node pairs simultaneously. Due to their high similarity, the labels of node pairs also have high confidence.

In summary, through the multi-source fusion selection strategy, our node pair labels can be defined as follows.
\begin{definition}\label{def:softlabel}
\textbf{Node Pair Labels.}  
Given the model prediction label $\hat{y}_{ij}$, the oracle's label $y_{ij}$, the model prediction accuracy $Acc$ and the predicted probability $p_{ij}$, the final label $\tilde{y}_{ij}$ is assigned as follows:
\label{eq11}
\begin{itemize}
    \item \textbf{High confidence region:} If $Acc \cdot p_{ij} > \alpha$, then $\tilde{y}_{ij} = \hat{y}_{ij}$;

    \item \textbf{Moderate confidence region:} If $\alpha > Acc \cdot p_{ij} > \gamma$, then
    \begin{itemize}
        \item If $y_{ij} = \hat{y}_{ij}$ or $y_{ij}^{\mathrm{twin}} = y_{ij}$, then $\tilde{y}_{ij} = y_{ij}$,
        \item If $y_{ij}^{\mathrm{twin}} = \hat{y}_{ij}$, then $\tilde{y}_{ij} = \hat{y}_{ij}$;
    \end{itemize}

    \item \textbf{Low confidence region:} If $\alpha > \gamma > Acc \cdot p_{ij}$, then $\tilde{y}_{ij} = y_{ij}$.
\end{itemize}
\end{definition}

\vspace{-8pt}
\subsection{Training}
In Stage 1, we incorporate existing network alignment models into our framework. The labeled pairs are used to train the model, which then provides predicted labels for node pairs with high confidence. In Stage 2, given the training set $\mathcal{V}_{train}$ and the batch size $b$, we iteratively select node pair $(v_i^s, v_j^t)$ with the maximum marginal gain \citep{nemhauser1978analysis}, and update the newly labeled node set $\mathcal{V}_l$, batch node set $\mathcal{V}_b$, and training node set $\mathcal{V}_{train}$. In Stage 3, we assign labels to the selected node pairs according to Eq. \eqref{eq11}. The algorithm of RANA can be found in Appendix \ref{alg}.

\section{Experiments}
In this section, we evaluate RANA through experiments designed to address the following key research questions:

\textbf{Q1}: Does RANA outperform other active learning baseline methods in noisy network alignment?

\textbf{Q2}: Is RANA robust to varying levels of structural noise and labeling noise?

\textbf{Q3}: Can the Noise-aware Selection Module effectively identify valuable node pairs for network alignment?

\textbf{Q4}: Does the Label Denoising Module significantly improve the labeling accuracy of the noisy oracle?

\vspace{-10pt}
\subsection{Experiment Setup}
\textbf{Dataset.} To validate the effectiveness of our method, we selected six publicly available real-world datasets with different types, scales, and attribute dimensions for the experiments. We paired them into three groups to form experimental datasets for network alignment. The detailed descriptions of these datasets are provided in Appendix \ref{datasets}.

\begin{table}[ht]
\caption{Statistics of Datasets.}
\vspace{0.5em}
\centering
\begin{tabular}{@{}ccccc@{}}
\toprule
\textbf{Dataset} & \textbf{Nodes} & \textbf{Edges} & \textbf{Groundtruth}  & \textbf{Attributes}  \\ \midrule
Douban Offline   & 1,118           & 1,511           & \multirow{2}{*}{1,118} & \multirow{2}{*}{538} \\
Douban Online    & 3,906           & 8,164           &                       &                      \\ \midrule
Allmovie         & 6011           & 124,709         & \multirow{2}{*}{5,174} & \multirow{2}{*}{14}  \\
Tmdb             & 5713           & 119,073         &                       &                      \\ \midrule
Facebook         & 1,043           & 4,734           & \multirow{2}{*}{1,043} & \multirow{2}{*}{0}   \\
Twitter          & 1,043           & 4,860           &                       &                      \\ \bottomrule
\end{tabular}
\end{table}

\textbf{Baseline.} We compared our proposed algorithm with several state-of-the-art and classic methods. PARROT \citep{zeng2023parrot} and NAME \citep{huynh2021network} are two advanced algorithms for network alignment. Attent \citep{zhou2021attent} is a novel active learning framework for network alignment, and TopMatchings \citep{malmi2017active} is a classic active learning method for network alignment. Entropy, Least Confident, Margin, and Random are four classic uncertainty sampling strategies. The summary and detailed description of these methods are provided in Appendix \ref{baseline}.

\textbf{Network Alignment Model.} To demonstrate the generality and effectiveness of RANA, we select three network alignment models as part of our framework to predict anchor links, including FINAL \citep{zhang2016final}, IsoRank \citep{singh2008global}, and PALE \citep{man2016predict}. FINAL aligns networks using both structural and attribute consistency through an efficient heuristic algorithm. IsoRank is a classical network alignment algorithm that relies solely on structural consistency and is suitable for small to medium-sized networks. PALE learns the local embedding representation of nodes and uses node features for alignment. Furthermore, we adopt PARROT as the base model in our framework, demonstrating the performance improvements our approach can achieve even on top of a strong SOTA backbone.

\textbf{Hyperparameters.} In RANA, we set the activation threshold $\theta = 0.05$ and the minimum acceptable threshold $\gamma = 0.01$. To ensure fairness, the network alignment models we use follow the parameter settings from the original papers. We repeat each method five times and the average test accuracy was reported to account for randomness.

\vspace{-10pt}
\subsection{Alignment Accuracy Analysis}

\begin{table*}[htbp]
\centering
\caption{Acc@1 on different datasets when the alignment model is FINAL.}
\vspace{0.5em}
\label{tab:table2}
\resizebox{\textwidth}{!}{ 
\begin{tabular}{c|ccccc|ccccc|ccccc}
\hline
\textbf{Dataset}       & \multicolumn{5}{c|}{\textbf{Douban Offline-Online}} & \multicolumn{5}{c|}{\textbf{Allmovie-Tmdb}} & \multicolumn{5}{c}{\textbf{Facebook-Twitter}} \\ \hline
Training Rate          & 0.1 & 0.2 & 0.3 & 0.4 & 0.5 & 0.1 & 0.2 & 0.3 & 0.4 & 0.5 & 0.1 & 0.2 & 0.3 & 0.4 & 0.5 \\ \hline
Random                 & 0.3828 & 0.5197 & 0.6547 & 0.6977 & 0.7558 & 0.7911 & 0.8777 & 0.9129 & 0.9387 & 0.9521 & 0.2685 & 0.4608 & 0.6749 & 0.7699 & 0.8408 \\ 
Least-Confident        & 0.3658 & 0.5295 & 0.6351 & 0.6762 & 0.7379 & 0.7986 & 0.8867 & 0.9242 & 0.9469 & 0.9595 & 0.2713 & 0.4986 & 0.7066 & 0.8121 & 0.8599 \\
Margin                 & 0.3658 & 0.5054 & 0.6351 & 0.6771 & 0.7379 & 0.7889 & 0.8879 & 0.9105 & 0.9353 & 0.9429 & 0.2685 & 0.4966 & 0.6942 & 0.7833 & 0.8428 \\
Entropy                & 0.3864 & 0.5179 & 0.6458 & 0.6887 & 0.7478 & 0.7891 & 0.8751 & 0.9192 & 0.9429 & 0.9517 & 0.2704 & 0.4727 & 0.6702 & 0.7651 & 0.8293 \\
TopMatchings           & 0.3846 & 0.5291 & 0.6592 & 0.6932 & 0.7487 & 0.7909 & 0.8751 & 0.9136 & 0.9382 & 0.9515 & 0.2483 & 0.4756 & 0.6807 & 0.7766 & 0.8369 \\ \hline
NAME                   & 0.3346 & 0.4955 & 0.6323 & 0.6725 & 0.7414 & 0.7605 & 0.8761 & 0.8987 & 0.9250 & 0.9525 & - & - & - & - & - \\
PARROT                 & 0.3485 & 0.5024 & 0.6743 & 0.7088 & 0.7668 & 0.7885 & 0.8665 & 0.9225 & 0.9387 & 0.9619 & 0.2694 & 0.4890 & 0.6424 & 0.7737 & 0.8456 \\ \hline
\textbf{RANA}          & \textbf{0.3936} & \textbf{0.5304} & \textbf{0.7048} & \textbf{0.7227} & \textbf{0.7835} & \textbf{0.8005} & \textbf{0.8896} & \textbf{0.9248} & \textbf{0.9474} & \textbf{0.9622} & \textbf{0.3241} & \textbf{0.5469} & \textbf{0.7315} & \textbf{0.8159} & \textbf{0.8611} \\ \hline
\end{tabular}
}
\end{table*}

\begin{table*}[htbp]
\centering
\vspace{0.5em}
\caption{Acc@1 on different datasets when the alignment models are PALE and IsoRank.}
\vspace{0.5em}
\label{tab:table3}
\resizebox{\textwidth}{!}{%
\begin{tabular}{c|ccccc|ccccc|ccccc}
\hline
\textbf{Alignment Model} & \multicolumn{5}{c|}{\textbf{PALE on Allmovie-Tmdb}} & \multicolumn{5}{c|}{\textbf{PALE on Facebook-Twitter}} & \multicolumn{5}{c}{\textbf{IsoRank on Facebook-Twitter}} \\ \hline
Training Rate            & 0.1 & 0.2 & 0.3 & 0.4 & 0.5 & 0.1 & 0.2 & 0.3 & 0.4 & 0.5 & 0.1 & 0.2 & 0.3 & 0.4 & 0.5 \\ \hline
TopMatchings             & 0.4884 & 0.6946 & 0.7590 & 0.7802 & 0.7868 & 0.1266 & 0.2972 & 0.4631 & 0.5580 & 0.6376 & 0.2800 & 0.4919 & 0.6290 & 0.7200 & 0.7710 \\
Entropy                  & 0.4733 & 0.7050 & 0.7592 & 0.7837 & 0.7938 & 0.1169 & 0.2780 & 0.4151 & 0.4910 & 0.6203 & 0.2694 & 0.4698 & 0.6174 & 0.7076 & 0.7651 \\
Margin                   & 0.4919 & 0.7051 & 0.7609 & 0.7797 & 0.7899 & 0.1352 & 0.3241 & 0.4775 & 0.5589 & 0.6462 & 0.3039 & 0.5101 & 0.6491 & 0.7335 & 0.7872 \\
Least-Confident          & 0.4907 & 0.7040 & 0.7607 & 0.7723 & 0.7922 & 0.1304 & 0.3164 & 0.4775 & 0.5628 & 0.6414 & 0.2943 & 0.5091 & 0.6635 & 0.7449 & 0.7967 \\
Random                   & 0.4785 & 0.7009 & 0.7600 & 0.7826 & 0.7928 & 0.1198 & 0.3145 & 0.4458 & 0.5417 & 0.6568 & 0.2694 & 0.4736 & 0.6242 & 0.7191 & 0.7728 \\ \hline
\textbf{RANA}            & \textbf{0.5058} & \textbf{0.7072} & \textbf{0.7613} & \textbf{0.7860} & \textbf{0.7976} & \textbf{0.1735} & \textbf{0.3701} & \textbf{0.5254} & \textbf{0.5954} & \textbf{0.6932} & \textbf{0.3663} & \textbf{0.5571} & \textbf{0.6741} & \textbf{0.7632} & \textbf{0.8303} \\ \hline
\end{tabular}
}
\end{table*}

\textbf{Network Alignment Accuracy on Different Datasets.} To answer \textbf{Q1}, we tested our active learning framework on three real world datasets and compared its alignment accuracy with that of other query node pair strategies and network alignment models. Additionally, we used three different network alignment models to validate the generality of RANA. For each model, we varied the proportion of training data from 0.1 to 0.5 and selected 100 node pairs for labeling via active learning. The experimental results are summarized in Table \ref{tab:table2} and Table \ref{tab:table3}.

From the results, we observe that RANA consistently outperforms all baseline methods across all datasets and alignment models on noisy network alignment task. For instance, in the Facebook-Twitter dataset, RANA achieved improvements of up to 5.28\%, 4.79\%, and 6.24\% over the best baseline at training data proportions of 0.1, 0.3, and 0.1, respectively, when using three different network alignment algorithms. Moreover, as the proportion of training data increased, RANA exhibited a clear upward trend in accuracy. This improvement highlights the effectiveness of the model-assisted labeling strategy, which becomes more reliable as the model's accuracy increases. RANA also outperforms existing network alignment algorithms, particularly when the amount of initial training data is limited. In such cases, alignment accuracy drops significantly, highlighting the substantial impact of labeling noise on alignment performance.

\begin{figure}[h]
\centering
\subfigure[Different query budget.]{\label{fig:figure2}
\includegraphics[width=0.45\columnwidth]{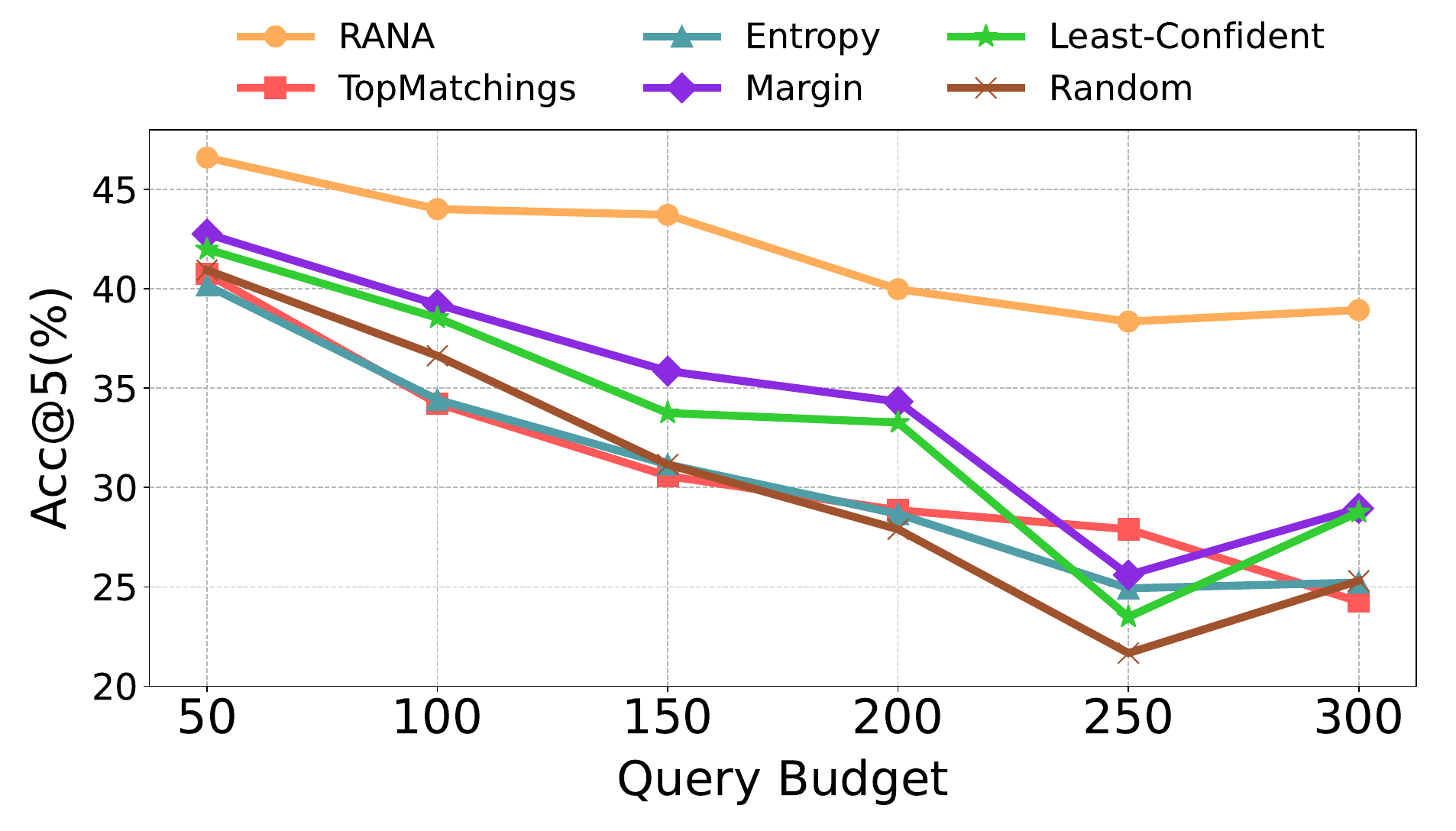}}
\subfigure[Different oracle accuracy.]{\label{fig:figure2b}
\includegraphics[width=0.45\columnwidth]{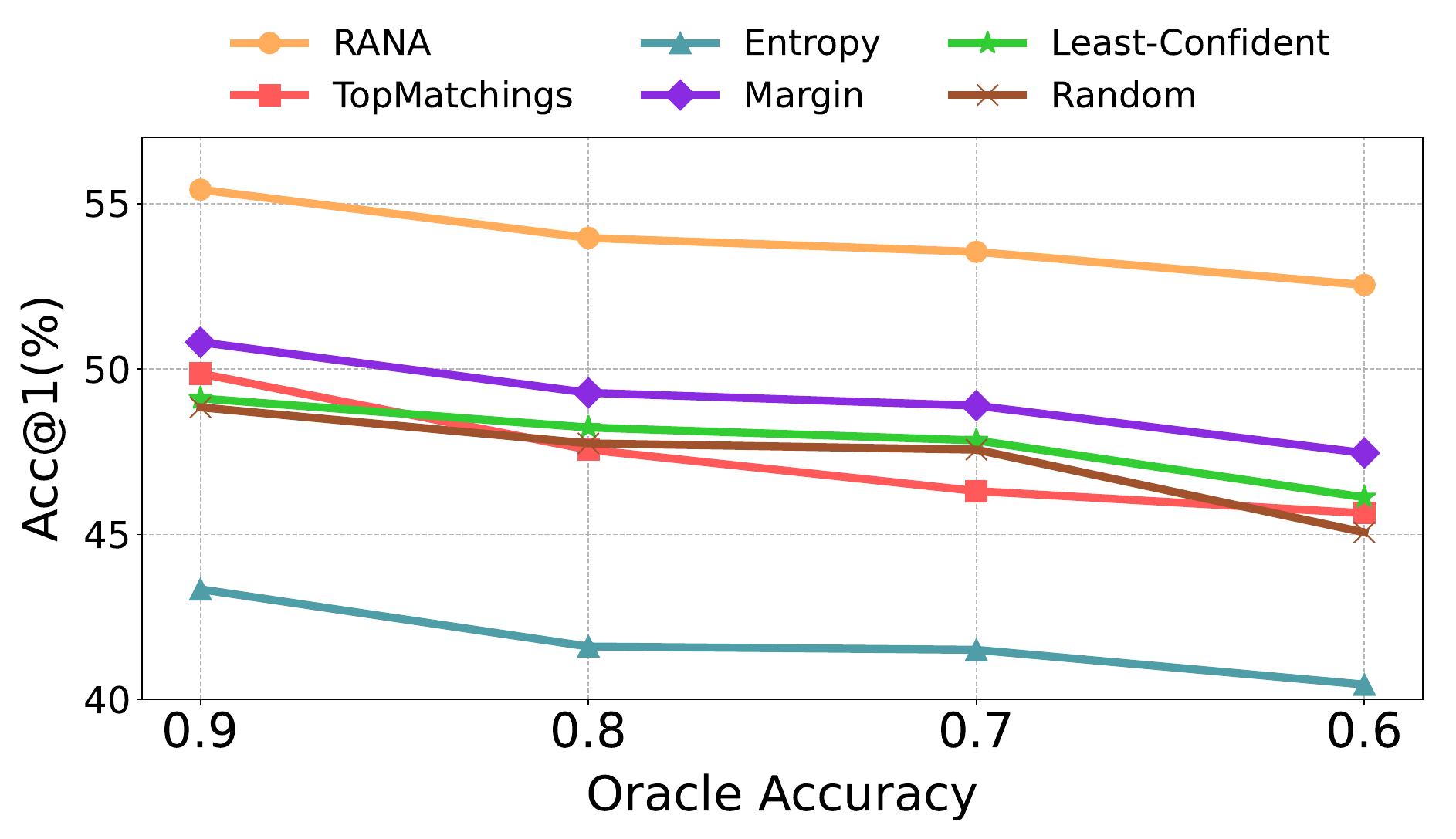}}
\caption{The alignment accuracy with different query budget and different oracle accuracy on the Facebook-Twitter dataset.}
\end{figure}

\textbf{Network alignment accuracy under different labeling budgets.} To validate the effectiveness of our framework in noisy environments, we evaluated the network alignment accuracy of various methods under different labeling budgets, as shown in Figure \ref{fig:figure2}. Using the Facebook-Twitter dataset, we varied the query budget $k$ from 50 to 300, with the oracle accuracy fixed at 0.8. The results demonstrate that as the query budget increases, our framework consistently outperforms other active learning methods. Initially, the alignment accuracy decreases as the query budget increases due to the introduction of labeling noise from the oracle. However, as more labeled data is acquired, the alignment accuracy gradually improves. RANA's alignment accuracy is also less sensitive to noise compared to other methods.

\subsection{Robustness Analysis}
The noise in noisy network alignment can be categorized into structural noise and labeling noise. Next, we analyze the robustness of our framework against structural and labeling noise through experiments to address \textbf{Q2}.

\begin{table}[ht]
\vspace{0.5em}
\caption{Acc@1 with different structural noise ratios.}
\vspace{0.5em}
\label{tab:table4}
\centering
\scriptsize
\setlength{\tabcolsep}{4pt}  
\renewcommand{\arraystretch}{1.3}
\begin{adjustbox}{width=\columnwidth}

\begin{tabular}{c|cccccc|c}
\hline
\makecell{\textbf{Noisy Edges} \\ \textbf{Ratios}} & 0 & 0.05 & 0.1 & 0.15 & 0.2 & 0.25 & \makecell{\textbf{Accuracy} \\ \textbf{Decrease}} \\
\hline
RANA            & 0.7055 & 0.7048 & 0.7034 & 0.7021 & 0.7006 & 0.6995 & \textbf{0.0060} \\
TopMatchings    & 0.6637 & 0.6619 & 0.6603 & 0.6592 & 0.6574 & 0.6533 & 0.0104 \\
Entropy         & 0.6619 & 0.6609 & 0.6601 & 0.6579 & 0.6500 & 0.6458 & 0.0161 \\
Margin          & 0.6391 & 0.6333 & 0.6324 & 0.6324 & 0.6315 & 0.6270 & 0.0121 \\
Least Confident & 0.6472 & 0.6423 & 0.6384 & 0.6319 & 0.6298 & 0.6288 & 0.0184 \\
\hline
\end{tabular}

\end{adjustbox}
\end{table}

\textbf{Network Alignment Accuracy Under Different Structural Noise Ratios.} To evaluate the robustness of our framework in environments with structural noise, we tested the alignment accuracy of various network alignment methods under different structural noise ratios, as shown in Table \ref{tab:table4}. Specifically, we introduced structural noise by adding a certain proportion of noisy edges to the original graph with the noisy edge ratio ranging from 5\% to 25\%. We use FINAL on the Douban dataset, with an initial training ratio of 0.3. The experimental results indicate that even as the proportion of noisy edges increases, RANA maintains a high level of alignment accuracy. In contrast, the alignment accuracy of other baseline methods declines significantly as the noisy edge ratio increases. These results demonstrate that our proposed framework exhibits stronger robustness and stability in handling structural noise, which is achieved through its noise-aware confidence and noise-aware maximization strategy.

\textbf{Network Alignment Accuracy Under Different Label Noise Ratios.} To validate the robustness of RANA to label noise, we evaluated its alignment accuracy against other baseline methods under different levels of label noise. The experimental results are shown in Figure \ref{fig:figure2b}. Specifically, we set the oracle labeling accuracy to range from 0.6 to 0.9, using PALE for network alignment on the Facebook-Twitter dataset, with an initial training data ratio of 0.3 and a query budget of 100. The results demonstrate that our framework consistently outperforms other methods across all label noise levels. Even as labeling accuracy decreases, RANA's alignment accuracy declines at a slower rate compared to baseline methods, indicating that our framework is less sensitive to noise label and exhibits strong robustness.

\vspace{-10pt}
\subsection{Ablation Study}
To address \textbf{Q3} and \textbf{Q4}, we conducted ablation experiments on the Noise-aware Selection Module and the Label Denoising Module to evaluate their respective contributions to improving network alignment performance. Also, we compare the performance of original model with its corresponding RANA version to evaluate how much accuracy improvement our framework can bring to network alignment models.

\begin{figure}[h]
\centering
\includegraphics[width=0.9\columnwidth]{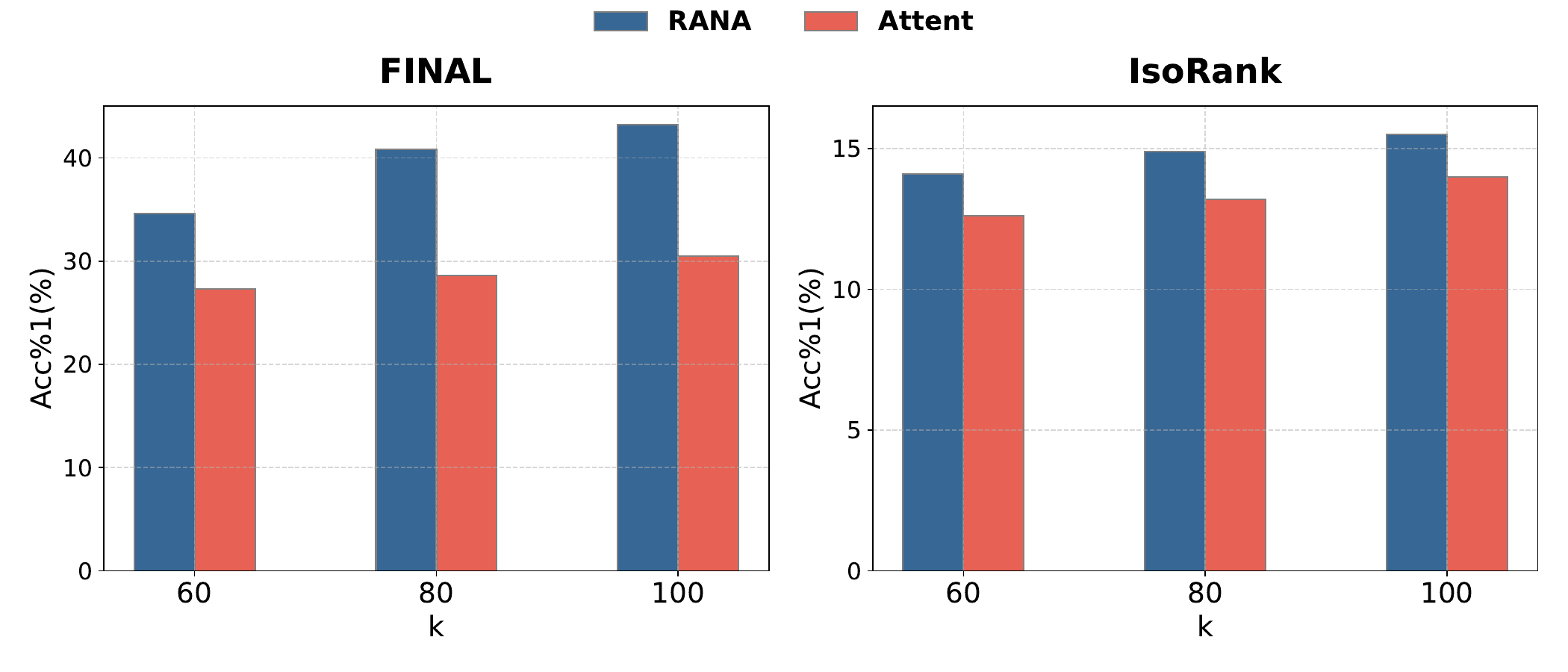}
\vspace{-0.5em}
\caption{Comparison of the node pair selection stage without noise.}
\vspace{1.5em}
\label{fig:figure3}
\end{figure}

\textbf{Node Pair Selection Stage.} In this stage, we set the oracle accuracy to 1 to exclude the influence of label noise and independently evaluate the node pair selection module in active learning. We compared our method with Attent, a state-of-the-art active learning strategy, and the results are summarized in Figure \ref{fig:figure3}. The results show that our node pair selection method significantly outperforms Attent, even in the absence of label noise. This demonstrates the effectiveness of the Noise-aware Selection Module in selecting informative and reliable node pairs. 

\vspace{-5pt}

\textbf{Node Pair Labeling Stage.} To verify the contributions of model-assisted labeling and twin node pairs labeling, we conducted a case study on the Douban and Facebook-Twitter datasets, as shown in Table \ref{tab:table6}. Specifically, we selected several node pairs with uncertain labels and compared the labels provided by the oracle, presenting their labels and related information. The results show that model-assisted labeling and twin node pair effectively provides more accurate labels in cases of labeling uncertainty. This indicates that the Label Denoising Module significantly alleviates the issue of inaccurate labels caused by labeling bias or noise, ultimately improving the overall accuracy of network alignment.

\begin{table}[ht]
\caption{Case study in Douban and Facebook-Twitter dataset.}
\vspace{0.5em}
\centering
\setlength{\tabcolsep}{4pt}
\label{tab:table6}
\footnotesize
\begin{tabular}{cccc}
\hline
\textbf{Node Pair} & \textbf{\begin{tabular}[c]{@{}c@{}}Model\& Oracle\\ Label\end{tabular}} & \textbf{\begin{tabular}[c]{@{}c@{}}Model's Predicted\\ Confidence\end{tabular}} & \textbf{\begin{tabular}[c]{@{}c@{}}Predicted\&\\ GroundTruth\end{tabular}} \\ \hline
\multicolumn{4}{c}{\textbf{Douban}}                               \\ \hline
(140, 3638)  & 1-1  & 0.7810  & 1-1 \\
(241, 1280)  & 1-1  & 0.7071  & 1-1 \\
(8, 429)  & 1-0 & 0.7952  & 1-1 \\ \hline
\multicolumn{4}{c}{\textbf{Facebook-Twitter}}  \\ \hline
(496, 496) & 0-1  & 0.0179    & 1-1 \\
(727, 727) & 1-0   & $9\times10^{-5}$  & 0-0 \\ \hline
\end{tabular}
\end{table}

\begin{figure}[h]
\centering
\includegraphics[width=0.9\columnwidth]{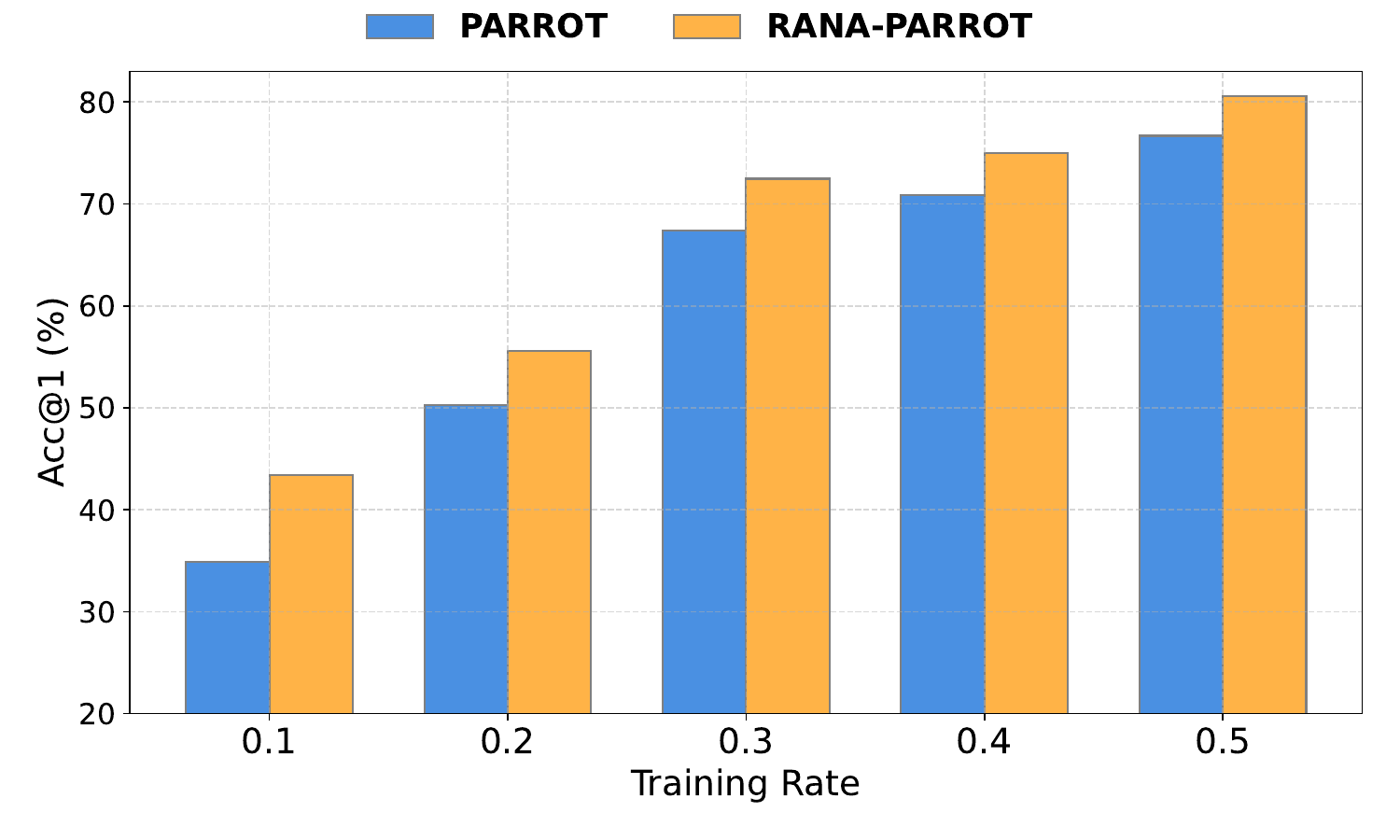}
\vspace{-0.5em}
\caption{Network alignment accuracy with and without RANA.}
\label{fig:figure4}
\end{figure}

\textbf{Performance improvements on network alignment models.} To evaluate the accuracy improvement our framework can bring to network alignment models, we conducted experiments on the SOTA network alignment model PARROT using the Douban dataset. The results are shown in Figure \ref{fig:figure4}. The results indicate that when the amount of training data is limited, our framework significantly mitigates the impact of noise and improves model accuracy by 8.57\%. Furthermore, even when the training data ratio increases to 0.5, the framework continues to yield a notable accuracy improvement of 3.87\%.

\vspace{-5pt}

\section{Conclusion}
In this paper, we address the issue of noisy network alignment. We propose a robust active learning framework including a Noise-aware Selection Module to actively identify informative and trustworthy node pairs, and a Label Denoising Module to mitigate the effect of label noise. Extensive experiments on multiple datasets demonstrate that our framework consistently outperforms existing methods while reducing annotation costs.



\begin{ack}
This work is supported by the National Key Research and Development Program of China (NO.2022YFB3102200), the National Natural Science Foundation of China (No.62402491), and the China Postdoctoral Science Foundation (No.2025M771524). 
\end{ack}



\bibliography{main}

\appendix
\section{Proof of Confidence Calculation}
\label{sec4.5}
Next, we will demonstrate how the confidence of node pairs is calculated. When the model's confidence is greater than the oracle's confidence or smaller than the minimum acceptable threshold, the node pair label is already determined, and we will not further elaborate on the confidence calculation in these two cases.

Based on the model's predicted label, the original node pair label obtained from the oracle, and twin node pair label, we consider the following three cases: 

\textbf{Case 1}: The model's predicted label matches the oracle's query label.

The oracle's and the model's predicted labels are independent, denoted as $\mathrm{C}^{orc}$ and $\mathrm{C}^{m}_{ij}$. If they match, then they must either both be correct or both be incorrect. The probabilities that both the oracle and the model predict correctly, or both predict incorrectly, are
\begin{equation}
\mathrm{C}^{orc}\cdot \mathrm{C}^{m}_{ij} 
\end{equation}
\begin{equation}
(1-\mathrm{C}^{orc})\cdot(1-\mathrm{C}^{m}_{ij})
\end{equation}

Thus, the reliability of the node pair in this case is
\begin{equation}
\frac{\mathrm {C}^{orc} \cdot \mathrm {C}^m_{ij}}{\mathrm {C}^{orc} \cdot \mathrm {C}^m_{ij}+(1-\mathrm {C}^{orc})\cdot (1-\mathrm {C}^m_{ij})} 
\end{equation}

\textbf{Case 2}: The model's predicted label does not match the oracle's query label, and the label obtained from querying the twin node pair matches the model's predicted label.

When the model's predicted label does not match the oracle's query label, it means that one of the predictions is correct and the other is incorrect, and the reliabilities of the two predictions are independent. The probability that the two predicted labels do not match is
\begin{equation}
1-\mathrm{C}^{orc}\cdot \mathrm{C}^m_{ij}
\end{equation}

To further determine the label of the original node pair, we also query the label of the twin node pair. If the twin node pair's label matches the model's predicted label, we consider the model's prediction to be correct, and the reliability of the node pair is
\begin{equation}
\frac{\mathrm {C}^m_{ij} \cdot(1-\mathrm {C}^{orc})}{1-\mathrm {C}^{orc}\cdot \mathrm {C}^m_{ij}}
\end{equation}

\textbf{Case 3}: The model's predicted label does not match the oracle's query label, and the label obtained from querying the twin node pair matches the oracle's query label.

If the twin node pair's label matches the oracle's query label, we consider the oracle's query label to be correct, and the reliability of the node pair is
\begin{equation}
\frac{\mathrm {C}^{orc} \cdot(1-\mathrm {C}^m_{ij})}{1-\mathrm {C}^{orc}\cdot \mathrm {C}^m_{ij}}
\end{equation}

\section{Algorithm of RANA}
\label{alg}
\begin{algorithm}
\small
\caption{Pipeline of RANA}\label{algorithm}
\KwIn{Initial labeled pairs set $\mathcal{V}_{train}$, oracle accuracy $\alpha$, query batch size $b$}
\KwOut{New labeled pairs set $\mathcal{V}_l$}
$\mathcal{V}_l = \emptyset$, $\mathcal{V}_b=\emptyset$, $\mathcal{V}_u = \mathcal{V}/\mathcal{V}_{train}$\;
\textbf{Stage 1: Model-assisted Labeling}\\
Train network alignment model with $\mathcal{V}_{train}$\;
Get prediction probability $p_{ij}$, labels $\hat{y}_{ij}$ and Accuracy $Acc$\;
\textbf{Stage 2: Node Pair Selection}\\
Calculate $\mathrm{C}_{ij}$ according to Eq. \eqref{eq4}\;
\For{1, 2, ..., b}{
    Select the most valuable pairs according to Eq. \eqref{objective}\;
}
\textbf{Stage 3: Node Pair Labeling}\\
Label pairs with $\hat{y}_{ij}$ if $Acc \cdot p_{ij} > \alpha$\;
\For{$(v_i^s, v_j^t)\in \mathcal{V}_b$}{
    Label $(v_i^s, v_j^t)$ according to Eq. \eqref{eq11}\;
    $\mathcal{V}_l=\mathcal{V}_l\cup \{(v_i^s, v_j^t)\}$\;
}
\end{algorithm}

\section{Experimental Details}
\subsection{Datasets Descriptions}\label{datasets}
\begin{itemize}
\item \textit{Douban Offline-Online} \citep{zhang2016final}: The Douban dataset is a social network dataset where nodes represent users and edges represent relationships between users. User attributes include location information and participation in offline activities, while edge attributes indicate whether two users are contacts or friends. This dataset contains 1118 anchor links.
\item \textit{Allmovie-Tmdb}: The Allmovie network is constructed based on data from Rotten Tomatoes. In this network, an edge exists between two movies if they share at least one actor. The Tmdb network is constructed in a similar manner to Allmovie. The alignment results between the networks consist of movie identifiers, with a total of 5176 anchor links.
\item \textit{Facebook-Twitter} \citep{cao2016asnets}: The Facebook and Twitter datasets are derived from Facebook and Twitter user accounts in Singapore. The authenticity of user identity pairs is partly based on their brief biography descriptions on their Twitter accounts. This dataset contains 1043 anchor links but lacks node attribute information.
\end{itemize}

\subsection{Baseline Details}\label{baseline}
\begin{itemize}
\item \textit{PARROT} \citep{zeng2023parrot}: PARROT addresses the network alignment problem by formulating it as a consistency-regularized optimal transport problem. To capture both structural and attribute information, PARROT constructs a position-aware transport cost matrix using random walk with restart on both separated and product graphs. Additionally, it introduces three consistency regularization terms to ensure alignment stability and robustness. To solve the resulting OT problem efficiently, PARROT applies a constrained proximal point method that decomposes the non-convex optimization into a series of convex subproblems. Among existing network alignment methods, PARROT consistently achieves the best overall performance across both plain and attributed networks.
\item \textit{NAME} \citep{huynh2021network}: NAME tackles the network alignment problem by learning multiple holistic embeddings for each node to fully capture diverse types of structural and semantic information. Specifically, NAME constructs three types of node embeddings: a skip-gram-based shallow embedding, a GCN-based deep embedding, and a global community-aware embedding. To integrate these embeddings, NAME employs a late-fusion mechanism with adaptive weighting, optimized through a data augmentation process using perturbed versions of the input networks. The final alignment matrix is computed as a weighted sum of the embedding-wise alignment scores. NAME has excellent robustness to noise and limited supervision.
\item \textit{Attent} \citep{zhou2021attent}: Attent quantifies the impact of candidate node pairs on the network alignment result by defining an influence function. The influence of node pair $q_{ij}$ is defined as $I(q_{ij}) = \frac{\partial f(x)}{\partial \mathbf{H}(i,j)}$, where $f(x)$ is the utility function regarding the alignment solution vector, and $\mathbf{H}(i,j)$ represents the alignment preference of node pair $q_{ij}$. The node influence is defined as the aggregation of the influence of node pairs between the node and all nodes in the target network. The Attent method selects the node with the highest influence for query, i.e., $\hat{v}=\underset{v \in \mathcal{V}_s}{\arg\max} I(q_v)$. Since the source code for Attent is not available, we only use it as a baseline to demonstrate the effectiveness of the node pair selection stage. 
\item \textit{TopMatchings} \citep{malmi2017active}: In the TopMatchings method, node determinism is quantified by constructing a weighted bipartite graph and calculating the top-$l$ maximum weight matching. The determinism of node $v$ is defined as $\text{Cert}(v) = \underset{u \in \mathcal{C}_v}{\max} P(M(v) = u | \mathcal{H})$, where $P(M(v) = u | \mathcal{H})$ represents the marginal distribution of node $v$ matching node $u$, based on the sampled matching set $\mathcal{H}$. If the distribution of node $v$ matching different nodes in the sample is fairly uniform, it indicates higher uncertainty in the best match for the node, i.e., lower determinism. TopMatchings selects the node with the lowest determinism for query, i.e., $\hat{v} = \underset{v \in \mathcal{V}_s }{\arg\min}\text{Cert}(v)$.
\item \textit{Entropy}: In the cross-network node similarity matrix $\mathbf{X}$, each row/column represents the similarity between a source network node and all nodes in the target network. If the similarity of a row/column is close to a uniform distribution, the entropy of the node's similarity distribution is large, indicating higher uncertainty in the alignment of that node. Therefore, we select the node with the maximum entropy for query, i.e., $\hat{v}=\underset{v \in \mathcal{V}_s}{\arg\max}\sum_{u \in \mathcal{V}_t}-\mathbf{X}(u, v) \log \mathbf{X}(u, v)$.
\item \textit{Least Confident}: An intuitive query strategy in active learning is to select the node with the least confidence in its prediction. We define the information of the query node as $I(v)=1-\underset{u}{\max}{\mathbf{X}(u, v)}$, and therefore select the node with the maximum information for query, i.e., $\hat{v} = \underset{v}{\max} I(v)$.
\item \textit{Margin}: Nodes with the smallest difference between the maximum and minimum predicted values are considered to have the highest alignment uncertainty. We define the information of the query node as $I(v) = \mathbf{X}(m, v) - \mathbf{X}(n, v)$, where $\mathbf{X}(m, v)$ and $\mathbf{X}(n, v)$ are the maximum and minimum values of $\mathbf{X}$, respectively. We select the node with the smallest difference for query, i.e., $\hat{v} = \underset{v}{\min} I(v)$.
\item \textit{Random}: Randomly select node pairs from the source network and target network for label querying.
\end{itemize}

\subsection{Evaluate Metrics}
We use \textit{Acc@k} \citep{zhang2016final} and MAP \citep{man2016predict} to evaluate the performance of our framework. \textit{Acc@k} indicates if a node’s true anchor match is present in a list of top-$k$ potential anchors. 
\begin{equation}
Acc@k=\frac{\sum_{u_{s}^{*} \in V_{s}} \mathds{1}_{S[u_{s}^{*}, u_{t}^{*}] \in R(u_{s}^{\star})}}{\#\{\text { ground truth anchor links }\}}
\end{equation}

We use MAP, which is Mean Average Precision, to measure ranking. 
\begin{equation}
\text{MAP}=mean(\frac{1}{ra})
\end{equation}
where $ra$ is the rank of true anchor target in the sorted list of anchor candidates.

\subsection{Time Complexity Analysis}
The running time of RANA can be divided into three parts: training the network alignment model, computing the influence and cleanliness scores, and selecting node pairs. The time complexity of the first part depends on the selected model. When we use PALE, the time complexity is O(kd|E|), where k is the number of iterations, d is the dimension of the embedding vector, and |E| is the number of edges in the network. The time complexity of calculating the influence score and cleanliness score is O(|E|) and O(mNd), where m is the average node degree and N is the number of anchor links. The time complexity of the third part is O(b). Therefore, the total time complexity of RANA is O(mNd), which mainly comes from the calculation of the cleanliness score.

\end{document}